\documentclass[10pt,twocolumn,letterpaper]{article}

\usepackage[pagenumbers]{cvpr} % To force page numbers, e.g. for an arXiv version

\definecolor{cvprblue}{rgb}{0.21,0.49,0.74}
\usepackage[pagebackref,breaklinks,colorlinks,allcolors=cvprblue]{hyperref}
\usepackage{algorithm}
\usepackage{algorithmic}
\usepackage{multirow}
\usepackage{colortbl}
\usepackage{tabularx}
\usepackage{color}
\usepackage[accsupp]{axessibility}  % Improves PDF readability for those with disabilities.
\usepackage{pifont}% http://ctan.org/pkg/pifont
\usepackage{makecell}
\usepackage{multicol}
\usepackage{lipsum}
\usepackage{threeparttable}
\usepackage{placeins}
\title{Robust Drone-View Geo-Localization via Content-Viewpoint Disentanglement}

\author{
Ke Li$^{1}$, Di Wang$^{1}$\thanks{Corresponding author}, Xiaowei Wang$^{1}$, Zhihong Wu$^{1}$, Yiming Zhang$^{2}$, Yifeng Wang$^{1}$, Quan Wang$^{1}$\\
$^{1}$Xidian University, Xi'an, China \quad $^{2}$University of California, San Diego, USA
}

\begin{document}
\maketitle
\begin{abstract}
Drone-view geo-localization (DVGL) aims to match images of the same geographic location captured from drone and satellite perspectives.
Despite recent advances, DVGL remains challenging due to significant appearance changes and spatial distortions caused by viewpoint variations. 
Existing methods typically assume that drone and satellite images can be directly aligned in a shared feature space via contrastive learning.
Nonetheless, this assumption overlooks the inherent conflicts induced by viewpoint discrepancies, resulting in extracted features containing inconsistent information that hinders precise localization.
In this study, we take a manifold learning perspective and model \textit{the feature space of cross-view images as a composite manifold jointly governed by content and viewpoint}. 
Building upon this insight, we propose \textbf{CVD}, a new DVGL framework that explicitly disentangles \textit{content} and \textit{viewpoint} factors. 
To promote effective disentanglement, we introduce two constraints: 
\textit{(i)} an intra-view independence constraint that encourages statistical independence between the two factors by minimizing their mutual information; and
\textit{(ii)} an inter-view reconstruction constraint that reconstructs each view by cross-combining \textit{content} and \textit{viewpoint} from paired images, ensuring factor-specific semantics are preserved.
As a plug-and-play module, CVD integrates seamlessly into existing DVGL pipelines and reduces inference latency.
Extensive experiments on University-1652 and SUES-200 show that CVD exhibits strong robustness and generalization across various scenarios, viewpoints and altitudes, with further evaluations on CVUSA and CVACT confirming consistent improvements.
\end{abstract}    
\section{Introduction}
\label{sec:intro}
\begin{figure}[!t]
\centering
\includegraphics[width=\linewidth]{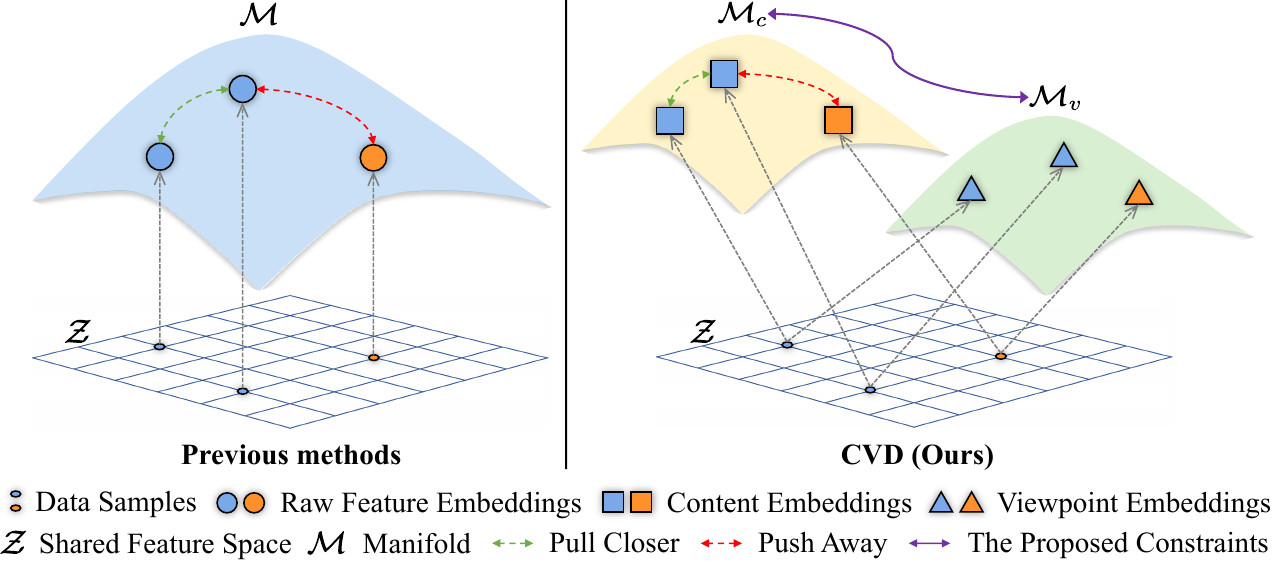} 
\vspace{-6mm}
\caption{Comparison between previous methods and our CVD. 
Left: Existing methods can be interpreted as operating on a single manifold $\mathcal{M}$, where contrastive objectives directly pull positive pairs closer and push negative pairs away.
Right: Our method learns disentangled representations by mapping inputs onto two submanifolds corresponding to \textit{content} $\mathcal{M}_c$ and \textit{viewpoint} $\mathcal{M}_v$.  
This separation is enforced via two constraints (see \cref{sec:independence} and \cref{sec:reconstruction}), promoting effective disentanglement and thereby enhancing cross-view matching performance.}
\label{fig:1}
\vspace{-3mm}
\end{figure}

The widespread deployment of drones and other intelligent systems has placed growing demands on navigation and localization technologies. 
To ensure safe and reliable operation, high-precision positioning services have become a fundamental requirement.
Drone-view geo-localization (DVGL), an onboard technique independent of external communication infrastructure, offers a promising solution by estimating absolute geospatial coordinates in the absence of conventional localization signals (\eg, GPS)~\cite{sun2023cross,dai2023vision,zhang2024aligning,lin2015learning}.
Given a drone image, the goal is to find a matching satellite image from a georeferenced database to infer the drone's location.
Most existing approaches formulate DVGL as an image retrieval task~\cite{sun2021multisensor,ye2024cross2}, training deep neural networks (DNNs) to learn visual similarity across different views.
However, the viewpoint disparity between drone and satellite imagery introduces severe spatial distortions and appearance variations, making robust matching inherently challenging.

Recent efforts aim to alleviate the viewpoint discrepancy between drone and satellite images.
A common strategy is to employ predefined geometric transformations to satellite images, such as polar transformation or orthorectification, aligning their spatial layout with that of the drone view.
However, the effectiveness of these methods relies on prior knowledge of the geometric relationship between the two views and may degrade when the drone image is not spatially centered within the satellite image~\cite{zhu2022transgeo}.

Despite recent advances in DVGL, many existing methods~\cite{liu2019lending,yang2021cross,zhu2022transgeo,deuser2023sample4geo,zhang2023cross} still follow a common training paradigm, \ie, directly applying contrastive learning to pull the features of positive pairs closer and push those of negative pairs away.
While various architectural or training refinements have been proposed, these methods largely overlook the semantic inconsistencies introduced by drastic viewpoint differences.
These inconsistencies disrupt the alignment between positive pairs and ultimately limit the drone-satellite matching performance.

In this paper, we revisit the DVGL task from a manifold learning perspective by modeling the feature space of cross-view images as a composite manifold jointly governed by \textit{content} and \textit{viewpoint} factors.
As illustrated in \cref{fig:1} (left), prior methods can be interpreted as learning representations on a single composite manifold $\mathcal{M}$ where both factors are entangled.
Such entanglement injects viewpoint-induced conflicts in the learned representations, undermining the robustness of contrastive alignment.
To address this limitation, we propose \textbf{CVD} (\textbf{C}ontent-\textbf{V}iewpoint \textbf{D}isentanglement), a general DVGL framework that explicitly factorizes the feature space into two submanifolds: content $\mathcal{M}_c$ and viewpoint $\mathcal{M}_v$ (see \cref{fig:1} (right)).
The \textit{content} encodes view-agnostic geo-structural information, while the \textit{viewpoint} captures view-specific appearance variations.
As shown in \cref{fig:2}, CVD adopts an \textbf{\textit{embed-disentangle-reconstruct}} paradigm: each image is first embedded into a shared feature space, then projected onto independent content and viewpoint submanifolds, and finally recombined via an image reconstruction task.
To facilitate the disentanglement, we impose two dedicated constraints: an intra-view independence constraint and an inter-view reconstruction constraint.
The former encourages statistical independence between \textit{content} and \textit{viewpoint} by minimizing their mutual information, while the latter preserves the intended semantics of each factor by reconstructing one view using the content from the drone (or satellite) image and the viewpoint from its satellite (or drone) counterpart.
In addition, we apply standard contrastive loss (\eg, InfoNCE) to align content representations of matched drone-satellite pairs.

To the best of our knowledge, CVD is the first DVGL framework to explicitly disentangle \textit{content} and \textit{viewpoint}.
Unlike approaches that require bespoke architectural redesigns, CVD integrates as a plug-and-play module into existing pipelines and lowering inference-time overhead.
Extensive experiments on four benchmarks, \ie University-1652, SUES-200, CVUSA and CVACT, demonstrate consistent gains over multiple baselines, improving robustness to viewpoint and altitude changes and generalization under scene shifts.
In summary, our contributions are as follows:
\begin{itemize}
\item 
We revisit the DVGL task from a manifold-learning perspective and propose CVD, the first framework that explicitly disentangles \textit{content} and \textit{viewpoint} to suppress viewpoint-induced conflicting information.
\item 
We introduce two constraints to facilitate disentanglement: an intra-view independence constraint that facilitates the independence between \textit{content} and \textit{viewpoint}, and an inter-view reconstruction constraint that ensures each factor preserves its intended semantics
\item 
CVD integrates seamlessly into existing pipelines, shortens inference time, and consistently improves cross-view matching performance, enabling efficient deployment in practice.
\item 
Extensive experiments show that CVD improves the robustness and generalization of DVGL pipelines across diverse scenarios, viewpoints, and altitudes.
\end{itemize}
\section{Related Work}
\label{sec:related_work}
\subsection{Visual-based Geo-Localization}\label{sec:related_vgl}
Visual-based geo-localization (VGL) has witnessed significant progress with the availability of large-scale geo-tagged datasets and advances in deep learning. 
Existing methods predominantly follow a Siamese-based framework and can be broadly categorized into three research directions: data augmentation strategies, architectural innovations, and feature representation learning.

\noindent \textbf{Data Augmentation Strategies.}
Data augmentation has become a widely adopted strategy in VGL tasks~\cite{vo2016localizing, rodrigues2021these, zhang2024benchmarking, li2024unleashing}.
To address cross-view misalignment, Liu~\textit{et al.}\cite{liu2019lending} incorporate camera orientation as auxiliary input, while SAFA~\cite{shi2019spatial} applies a polar transformation to align aerial and ground-level panoramas.
CVGlobal~\cite{ye2024cross} introduces a panoramic BEV transformation based on the ground-plane assumption and geometric constraints, effectively reducing the gap between street panoramas and satellite imagery.
Similarly, Video2BEV~\cite{ju2025video2bev} transforms drone videos into BEV representations, facilitating better alignment with satellite imagery.
More recently, training-aware data sampling has emerged as a complementary strategy.
Sample4Geo~\cite{deuser2023sample4geo} introduces two curriculum-driven strategies: one leveraging geographically adjacent samples for easier early-stage alignment, and another mining hard negatives to refine the decision boundary.
DenseUAV~\cite{dai2023vision} integrates metric learning with mutual supervision, effectively reducing modality discrepancy and improving feature discriminability.
Game4Loc~\cite{ji2025game4loc} proposes a mutual-exclusion sampling mechanism that enforces strict decorrelation between positive and negative pairs, thereby enhancing contrastive supervision in cross-view matching.

\noindent \textbf{Backbone Innovations.}
In parallel, a line of work focuses on designing more powerful visual backbones to enhance localization performance~\cite{rodrigues2022global,zhu2023modern}.
For example, L2LTR~\cite{yang2021cross} exploits self-attention to model long-range dependencies, effectively reducing visual ambiguity in cross-view.
RK-Net~\cite{lin2022joint} introduces a lightweight unit-difference attention module that enables joint learning of dense features and salient keypoints, without requiring additional annotations.
SAIG~\cite{zhu2023simple} proposes an efficient backbone tailored for VGL by replacing the MLP blocks in standard Transformers with spatially-aware mixing layers and low-dimensional projections, yielding a more compact and structured representation.

\noindent \textbf{Feature Representation Learning.}
Many studies focus on learning more effective visual representations to enhance cross-view matching~\cite{zhu2021vigor,mi2024congeo,zhu2023sues,zhai2017predicting,zeng2022geo,ti2024tirsaa}.
A common strategy involves refining alignment mechanisms between views.
For instance, Shi \textit{et al.}\cite{shi2020looking} proposed a dynamic similarity matching network to estimate directional alignment, thereby reducing cross-view discrepancies.
FSRA~\cite{dai2021transformer} leverages transformer-based heatmaps to perform region-level alignment, while LPN~\cite{wang2021each} incorporates contextual cues via a square-ring partitioning strategy to improve part-based representations.
SDPL~\cite{chen2024sdpl} builds upon LPN by introducing a shifting-fusion mechanism to generate multiple complementary part sets, which are then adaptively aggregated to enhance robustness against spatial shifts and scale variations.
Several methods also incorporate inductive priors modeling to enhance feature expressiveness.
FRGeo~\cite{zhang2024aligning} enhances cross-view alignment by explicitly recombining spatial features to reduce geometric ambiguities.
TransGeo~\cite{zhu2022transgeo} adopts a non-uniform cropping strategy that discards low-information regions while reallocating resolution to semantically salient areas, enhancing accuracy without increasing computational cost.
MCCG~\cite{shen2023mccg} enriches feature diversity by jointly modeling spatial and channel-wise attentions.

Although prior methods perform well across various scenarios, many rely on auxiliary components that increase inference-time overhead.
In contrast, we propose a plug-and-play training paradigm that explicitly disentangles \textit{content} and \textit{viewpoint} from raw feature representations, thereby effectively enhancing cross-view correspondence and reducing inference latency and computational cost.

\subsection{Disentangled Representation Learning}\label{sec:related_drl}
Disentangled representation learning (DRL) aims to learn representations that identify and disentangle the underlying factors hidden in observable data \cite{wang2024disentangled}.
Owing to the resulting interpretability, controllability, and robustness, it has seen broad adoption in computer vision \cite{li2025fd2,ruan2022adaptive,wang2016multimodal,chen2024disendreamer,cheng2024disentangled}, natural language processing \cite{zhao2024disentangled,bao2019generating,cheng2020improving}, recommender systems \cite{ma2019learning}, and graph learning \cite{wang2020disentangled,zhang2023dyted,wang2022disencite}, with gains on many downstream tasks.
For example, Zou~\textit{et al.}~\cite{zou2020joint} tackle cross-domain person re-identification by jointly disentangling ID-related and ID-unrelated subspaces and restricting adaptation to the former, thereby improving transferability.
DisCo~\cite{Wang_2024_CVPR} introduces a disentangled-control architecture that separates subject, background, and pose, enabling compositional and generalizable dance video synthesis. 
Wang~\textit{et al.}~\cite{wang2024towards} present a frequency-domain disentanglement framework for UAV object detection that employs two learnable filters to isolate domain-invariant from domain-specific spectra, leading to stronger domain generalization.

Recent studies have begun to incorporate DRL into VGL tasks.
GeoDTR~\cite{zhang2023cross} introduces a geometry-aware layout extractor to separate geometric cues from raw appearance features, thereby improving cross-view localization.
However, it leaves unaddressed the viewpoint-induced conflicts persisting in both appearance and layout representations, thereby hindering cross-view correspondence.
In this work, we explicitly disentangle \textit{content} from \textit{viewpoint}, and employ independence and reconstruction constraints to suppress such conflicts, yielding a cleaner, view-agnostic content representation.
\section{Methods}\label{sec:method}
\subsection{Problem Formulation}\label{sec:problem}
Considering a set of image pairs $\{(\mathcal{I}_i^{\mathrm{d}}, \mathcal{I}_i^{\mathrm{s}})\}_{i=1}^N$, where superscripts $\mathrm{d}$ and $\mathrm{s}$ denote drone and satellite images, respectively, and $N$ is the number of pairs. Each pair depicts the same geographic location. In the DVGL task, given a drone image $\mathcal{I}_d^{\mathrm{d}}$ with index $d$, the objective is to retrieve its best-matching satellite image $\mathcal{I}_b^{\mathrm{s}}$ from the georeferenced database, where $b \in \{1, \ldots, N\}$.

Most existing methods rely on learning a representation function $f(\cdot)$ that embeds images from different viewpoints into a shared feature space, allowing matching pairs to be identified through feature distance minimization. However, such representations inevitably retain view-specific conflicts, which hinder cross-view semantic alignment and degrade matching performance.

For effective comparison between cross-view images, we aim to modulate raw representations by explicitly disentangling two factors: \textit{content} and \textit{viewpoint}. 
Concretely, we model the feature space $\mathcal{Z}$ as a representation of \textit{composite manifold} $\mathcal{M}$ structured by two independent submanifolds, $\mathcal{M}_c$ and $\mathcal{M}_v$, corresponding to \textit{content} and \textit{viewpoint}, respectively.
Each feature representation is viewed as a sample from two latent random variables, $C \sim p(C)$ and $V \sim p(V)$, defined over $\mathcal{M}_c$ and $\mathcal{M}_v$. These are composed via a function $f: \mathcal{M}_c \times \mathcal{M}_v \rightarrow \mathcal{Z}$ that maps both factors to a point in the feature space.
Assuming statistical independence between the factors, \ie, $p(C, V) = p(C)\,p(V)$, the resulting distribution over $\mathcal{Z}$ is given by the push-forward measure $f_{\#}(p(C) \times p(V))$. This assumption is well-aligned with the DVGL task, where identical scenes may be observed under diverse perspectives.

To realize the above formulation, we design manifold encoders that decompose each image $\mathcal{I}^\mathrm{u}$ into two distinct representations: a content embedding $f_c(\mathcal{I}^\mathrm{u}) \in \mathcal{M}_c$ and a viewpoint embedding $f_v(\mathcal{I}^\mathrm{u}) \in \mathcal{M}_v$. 
Cross-view matching is subsequently performed in the $\mathcal{M}_c$ by retrieving the nearest neighbor of a drone image  $\mathcal{I}_q^{\mathrm{d}}$ via its content embedding:
\begin{equation}
    b = \underset{i \in\{1, \ldots, N\}}{\arg \min } d(f_c(\mathcal{I}_d^{\mathrm{d}}), f_c(\mathcal{I}_i^{\mathrm{s}})),
\end{equation}
where $d(\cdot, \cdot)$ denotes a distance metric.
For notation compactness, we will use superscript $u$\footnote[2]{We adopt this convention throughout the paper.} for cases that apply to both drone ($\mathrm{d}$) and satellite ($\mathrm{s}$) views.

\subsection{Proposed Methodology}
\label{sec:methodology}
\begin{figure}[!t]
\centering
\includegraphics[width=\linewidth]{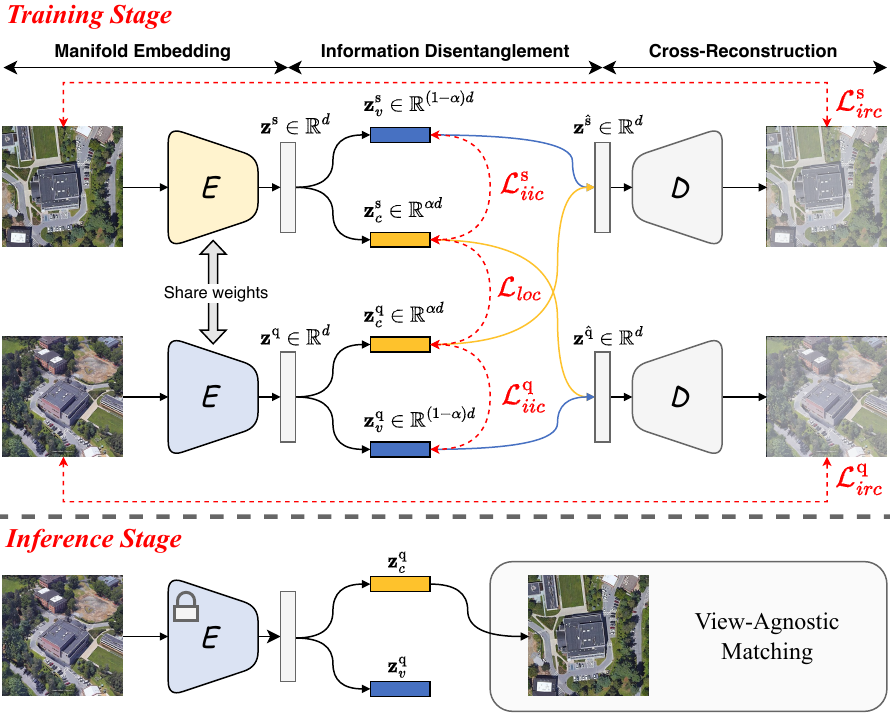} 
\vspace{-6mm}
\caption{Overview of the proposed CVD.}
\vspace{-4mm}
\label{fig:2}
\end{figure}

As illustrated in \cref{fig:2}, \textbf{CVD} adopts a Siamese architecture consisting of two symmetric branches for the drone and satellite views. 
Each branch comprises three sequential components: manifold embedding, information disentanglement, and cross-reconstruction.

\noindent \textbf{Manifold Embedding.} Each input image $\mathcal{I}^\mathrm{u}$ ($u \in \{\mathrm{d}, \mathrm{s}\}$) is first processed by manifold encoders $E$, yielding a raw $d$-dimensional feature representations $\mathbf{z}^\mathrm{u} = E(\mathcal{I}^\mathrm{u}) \in \mathbb{R}^d$. Owing to the nature of DNN encoders, the distribution of $\mathbf{z}^\mathrm{u}$ can be viewed as residing on a \textit{composite manifold} jointly governed by \textit{content} and \textit{viewpoint} information.
Since our primary focus is on the training paradigm, we adopt the same DNN encoders as those used in the respective baselines to ensure fair and consistent comparisons.

\noindent \textbf{Information Disentanglement.} Once we obtain the feature representation $\mathbf{z}^\mathrm{u}$, we project it into two statistically independent components using two parallel $3 \times 3$ convolutional layers with a channel ratio of $\alpha$.
One is used to represent the content embedding, denoted as $\mathbf{z}_c^\mathrm{u} = f_c(\mathbf{z}^\mathrm{u}) \in \mathbb{R}^{\alpha d}$, which provides view-agnostic scene structure.
The other is employed for the viewpoint embedding, denoted as $\mathbf{z}_v^\mathrm{u} = f_v(\mathbf{z}^\mathrm{u}) \in \mathbb{R}^{(1-\alpha)d}$, capturing view-specific attributes.
To promote effective disentanglement, we introduce an \textit{intra-view independence constraint} (\cref{sec:independence}) that minimizes the mutual information $\mathrm{MI}(Z_c^\mathrm{u}; Z_v^\mathrm{u})$, where $Z^\mathrm{u}$ denotes the random variables of $\mathbf{z}^\mathrm{u}$, thereby encouraging statistical independence between \textit{content} and \textit{viewpoint} factors.

\noindent \textbf{Cross-Reconstruction.}  
While the independence constraint enhances factor separation, it may inadvertently lead to degenerate solutions or information loss. 
To mitigate this, we introduce an \textit{inter-view reconstruction constraint} (\cref{sec:reconstruction}) that encourages each factor to retain its intended information through cross-view image reconstruction.
Specifically, we train two decoders, $D^\mathrm{d}$ and $D^\mathrm{s}$, to perform bidirectional reconstruction between paired views by swapping content and viewpoint embeddings, \ie, reconstructing each image using its own viewpoint and the content of the other.
By enforcing accurate reconstruction from these hybrid embeddings, the model is incentivized to encode factor-specific information in each representation.
This cross-view supervision not only prevents information collapse but also reinforces disentanglement.
In the following, we describe the two specific constraints in CVD.

\subsection{Intra-view Independence Constraint}
\label{sec:independence}
To effectively disentangle \textit{content} and \textit{viewpoint} factors, we introduce an intra-view independence constraint that aims to minimize the statistical dependence between the two embeddings.
Motivated by the principle that mutual information provides a fundamental measure of statistical dependence, we seek to encourage independent factorization by minimizing it between \textit{content} and \textit{viewpoint}.

Formally, the mutual information between $Z_c^\mathrm{u}$ and $Z_v^\mathrm{u}$ is defined as the Kullback-Leibler (KL) divergence between their joint distribution and the product of marginals:
\begin{equation}
\mathrm{MI}(Z_c^\mathrm{u}; Z_v^\mathrm{u}) = \mathcal{D}_{\mathrm{KL}}\left(p(\mathbf{z}_c^u, \mathbf{z}_v^u) \;\middle\|\; p(\mathbf{z}_c^u)\,p(\mathbf{z}_v^u)\right).
\end{equation}
However, direct computation or optimization of mutual information is notoriously intractable in high-dimensional feature spaces due to the need for accurate estimation of joint and marginal densities. To circumvent this issue, we adopt the \textit{Sliced Wasserstein Distance} (SWD) as a geometry-aware and sample-efficient proxy to promote independence. Specifically, we minimize the SWD between the empirical joint distribution $p(\mathbf{z}_c^u, \mathbf{z}_v^u)$ and the product of its marginals $p(\mathbf{z}_c^u)\,p(\mathbf{z}_v^u)$, which can be represented as:
\begin{equation}
\mathcal{L}_{\text{iic}}^\mathrm{u} = \mathcal{SW}_2\left(p(\mathbf{z}_c^u, \mathbf{z}_v^u),\; p(\mathbf{z}_c^u) \otimes p(\mathbf{z}_v^u)\right),
\label{eq:iicloss}
\end{equation}
where $\mathcal{SW}_2(\cdot,\cdot)$ denotes the Sliced Wasserstein-2 distance. 
We apply this constraint independently to both views, yielding $\mathcal{L}_{\text{iic}}^{\mathrm{d}}$ and $\mathcal{L}_{\text{iic}}^{\mathrm{s}}$.  
This constraint drives the separation of view-agnostic scene structure from view-specific attributes in a computationally tractable manner.

\subsection{Inter-view Reconstruction Constraint}
\label{sec:reconstruction}
While the independence constraint promotes factor separation, it does not guarantee that each embedding retains the essential factor-specific information.
In particular, relying solely on independence may lead to trivial solutions where either the content or the viewpoint embedding becomes uninformative.

To address this, we introduce an \textit{inter-view reconstruction constraint} that enforces information retention through cross-view image reconstruction. 
Specifically, we deploy decoders $D$ that reconstruct each image from a hybrid embedding composed of \textit{content} from one view with the \textit{viewpoint} from the other, which is denoted as:
\begin{equation}
\hat{\mathcal{I}}^{\mathrm{d}} = D^{\mathrm{d}}(\mathbf{z}_c^{\mathrm{s}}, \mathbf{z}_v^{\mathrm{d}}), \quad 
\hat{\mathcal{I}}^{\mathrm{s}} = D^{\mathrm{s}}(\mathbf{z}_c^{\mathrm{d}}, \mathbf{z}_v^{\mathrm{s}}).
\label{eq:ircloss}
\end{equation}
This constraint ensures that both $\mathbf{z}^\mathrm{u}_c$ and $\mathbf{z}^\mathrm{u}_v$ preserve distinct and sufficient information, thereby preventing representational collapse.
Notably, the reconstruction is conditioned on the \textit{viewpoint}, which governs spatial and geometric layout, while the \textit{content} determines underlying scene structure. 
The reconstruction loss is defined as:
\begin{equation}
\mathcal{L}^\mathrm{u}_{\text{irc}} = \| \mathcal{I}^\mathrm{u} - \hat{\mathcal{I}}^\mathrm{u} \|_2^2.
\end{equation}
where $\| \cdot \|_2^2$ denotes the mean squared error (MSE) between the original and reconstructed images. 
This loss is also applied to both views, resulting in $\mathcal{L}_{\text{irc}}^{\mathrm{d}}$ and $\mathcal{L}_{\text{irc}}^{\mathrm{s}}$, which ensure that the disentangled features preserve the necessary information to reconstruct their cross-view counterparts.

\subsection{Training Objective}
\label{sec:training_objective}
Following prior works \cite{deuser2023sample4geo,mi2024congeo}, we employ the standard InfoNCE loss for view-agnostic content consistency across views, denoted as $\mathcal{L}_{\text{loc}}$, which encourages \textit{content} of matched drone-satellite pairs to be close while pushing away mismatched pairs, and is defined as:
\begin{equation}
\mathcal{L}_{\text{loc}} = -\log \frac{\exp(\mathbf{z}_c^{\mathrm{d}} \cdot \mathbf{z}_c^{\mathrm{s}} / \tau)}{\sum_{i=1}^N \exp(\mathbf{z}_c^{\mathrm{d}} \cdot \mathbf{z}_c^{\mathrm{i}} / \tau)},
\label{eq:locloss}
\end{equation}
where $\tau$ is a temperature parameter that controls the sharpness of the similarity distribution.
The overall training objective for CVD combines three losses: (1) an intra-view independence loss $\mathcal{L}^\mathrm{u}_{\text{iic}}$ promotes \textit{content} and \textit{viewpoint} independence, (2) an inter-view reconstruction loss $\mathcal{L}^\mathrm{u}_{\text{irc}}$ to ensure intended information preservation, and (3) a cross-view localization loss $\mathcal{L}_{\text{loc}}$ for discriminative alignment, which can be expressed as:
\begin{equation}
\mathcal{L}_{\text{total}} = 
\lambda_1 ( \tfrac{1}{2} \mathcal{L}^{\mathrm{d}}_{\text{iic}} + \tfrac{1}{2} \mathcal{L}^{\mathrm{s}}_{\text{iic}} ) +
\lambda_2 ( \tfrac{1}{2} \mathcal{L}^{\mathrm{d}}_{\text{irc}} + \tfrac{1}{2} \mathcal{L}^{\mathrm{s}}_{\text{irc}} ) + 
\mathcal{L}_{\text{loc}},
\label{eq:total_loss}
\end{equation}
where $\lambda_{1}$ and $\lambda_{2}$ are two loss-balancing hyperparameters.
% The algorithm of CVD is summarized in \cref{algo:cvd}.
\section{Experiments}\label{sec:experiments}
\subsection{Settings}
\noindent \textbf{Datasets.}
We evaluate our method on four representative CVGL benchmarks:
\textbf{University-1652} \cite{zheng2020university} comprises images from 1,652 university campuses captured in ground, drone, and satellite views. In our experiments, we adopt the drone-satellite setting, using 701 campuses for training, 701 for testing, along with 250 distractor samples.
\textbf{SUES-200} \cite{zhu2023sues} consists of real-world drone and satellite imagery from 200 scenes across four altitudes (150-300m), with 120 scenes for training and 80 for testing.
Together, University-1652 and SUES-200 span diverse scene types, flight altitudes, and viewing directions, providing a rigorous testbed to assess the robustness and generalization of our approach.
To further validate effectiveness, we additionally evaluate on the \textbf{CVUSA} \cite{workman2015wide} and \textbf{CVACT\_val} \cite{liu2019lending} datasets (ground $\rightarrow$ satellite), which provide 35,532 aligned training pairs each; the former contains 8,884 test queries, and the latter offers a test set of the same size.

\noindent \textbf{Evaluation metrics.}
We adopt five retrieval metrics, including Average Precision (AP), Recall@K (K=1,5,10), and Recall@1\%, to evaluate cross-view matching performance. 
Definitions of these metrics are provided in the Appendix.

\noindent \textbf{Baselines.}
To evaluate the effectiveness of our method, we compare it against several SOTA methods, including LPN~\cite{wang2021each}, FSRA~\cite{dai2021transformer}, TransGeo~\cite{zhu2022transgeo}, MCCG~\cite{shen2023mccg}, Sample4Geo~\cite{deuser2023sample4geo}, SDPL~\cite{chen2024sdpl}, Game4Loc~\cite{ji2025game4loc} and GeoDTR~\cite{zhang2023cross}.

\noindent \textbf{Implementation details.} 
The hyperparameters of the total training loss are fixed across all experiments, with $\lambda_1 = 10$, and $\lambda_2 = 0.2$. 
The temperature parameter $\tau$ is set to 0.05.
To ensure fair comparisons, we strictly follow the original training configurations of each baseline, including optimizer type, learning rate schedule, \etc, without any additional tuning.
All experiments are conducted in PyTorch on an NVIDIA 4090 GPU.
Each experiment is repeated three times with different random seeds, and the mean results are reported to ensure statistical reliability.

\begin{table*}[!t]
\centering
\caption{Comparison of baselines and their CVD-enhanced counterparts (marked with $\dagger$) on the University-1652 dataset.}
\label{tab:univ1652}
\vspace{-3mm}
\resizebox{\textwidth}{!}{
\begin{tabular}{lcccccccccccc}
    \toprule
    \multirow{2}{*}{Method} & \multirow{2}{*}{Image Size} & \multirow{2}{*}{Backbone} & \multicolumn{5}{c}{\textbf{Drone $\rightarrow$ Satellite}} & \multicolumn{5}{c}{\textbf{Satellite $\rightarrow$ Drone}} \\
    \cmidrule(lr){4-8} \cmidrule(lr){9-13}
    & & & AP & R@1 & R@5 & R@10 & R@1\% & AP & R@1 & R@5 & R@10 & R@1\% \\
    \midrule
    LPN          & 256$\times$256 & ResNet50   & 77.26 & 73.87 & 88.84 & 92.58 & 93.01 & 73.55 & 85.28 & 89.27 & 91.15 & 98.29 \\
    LPN$\dagger$ & 256$\times$256 & ResNet50   & \textbf{78.99} & \textbf{75.78} & \textbf{89.96} & \textbf{93.45} & \textbf{93.81} & \textbf{74.89} & \textbf{85.88} & \textbf{90.73} & \textbf{92.58} & \textbf{99.00} \\
    \hline
    FSRA                  & 256$\times$256 & ViT-S   & 84.24 & 81.62 & 93.06 & 95.19 & 95.43 & 80.99 & 87.87 & 90.87 & 92.87 & 98.43 \\
    {FSRA$\dagger$}         & 256$\times$256 & ViT-S   & \textbf{85.53} & \textbf{83.32} & \textbf{94.38} & \textbf{96.12} & \textbf{96.56} & \textbf{82.28} & \textbf{88.39} & \textbf{91.91} & \textbf{93.86} & \textbf{99.17} \\
    \hline
    SDPL                  & 512$\times$512 & ResNet50   & 86.45 & 84.13 & 94.36 & 96.45 & 96.72 & 82.17 & 89.44 & 92.58 & 93.58 & 99.29 \\
    SDPL$\dagger$         & 512$\times$512 & ResNet50   & \textbf{87.24} & \textbf{84.98} & \textbf{95.21} & \textbf{96.91} & \textbf{97.19} & \textbf{82.98} & \textbf{89.83} & \textbf{93.15} & \textbf{94.58} & \textbf{99.43} \\
    \hline
    MCCG                  & 256$\times$256 & ConvNeXt-T   & 90.63 & 88.92 & 96.44 & 97.63 & 97.77 & 88.73 & 93.15 & 95.72 & 96.72 & 99.57 \\
    MCCG$\dagger$         & 256$\times$256 & ConvNeXt-T   & \textbf{92.19} & \textbf{90.67} & \textbf{97.65} & \textbf{98.55} & \textbf{98.67} & \textbf{89.63} & \textbf{93.65} & \textbf{96.71} & \textbf{97.84} & \textbf{99.68} \\
    \hline
    Sample4Geo            & 384$\times$384 & ConvNeXt-B & 93.56 & 92.36 & 97.64 & 98.26 & 98.36 & 91.64 & 94.72 & 97.00 & 97.43 & 99.43  \\
    Sample4Geo$\dagger$   & 384$\times$384 & ConvNeXt-B & \textbf{94.78} & \textbf{93.73} & \textbf{98.56} & \textbf{98.90} & \textbf{98.95} & \textbf{92.48} & \textbf{95.26} & \textbf{97.76} & \textbf{98.47} & \textbf{99.55}  \\
    \hline
    Game4Loc              & 384$\times$384 & ViT-B      & 92.56 & 91.32 & 96.56 & 97.33 & 97.41 & 90.83 & 94.43 & 95.72 & 96.57 & 98.71  \\
    Game4Loc$\dagger$     & 384$\times$384 & ViT-B      & \textbf{93.95} & \textbf{92.94} & \textbf{97.59} & \textbf{98.28} & \textbf{98.32} & \textbf{91.92} & \textbf{94.86} & \textbf{96.71} & \textbf{97.14} & \textbf{99.57}  \\
    \bottomrule
\end{tabular}}
\vspace{-1mm}
\end{table*}

\begin{table*}[!t]
\centering
\caption{Comparison of baselines and their CVD-enhanced counterparts (marked with $\dagger$) on the SUES-200 dataset.}
\label{tab:sues200}
\vspace{-3mm}
\setlength{\tabcolsep}{4pt}
\resizebox{\linewidth}{!}{
\begin{tabular}{lcccccccccccccccc}
    \toprule
    \multirow{3}{*}{Method} & \multicolumn{8}{c}{\textbf{Drone $\rightarrow$ Satellite}} & \multicolumn{8}{c}{\textbf{Satellite $\rightarrow$ Drone}} \\
    \cmidrule(lr){2-9} \cmidrule(lr){10-17}
    & \multicolumn{2}{c}{150m} & \multicolumn{2}{c}{200m} & \multicolumn{2}{c}{250m} & \multicolumn{2}{c}{300m} & \multicolumn{2}{c}{150m} & \multicolumn{2}{c}{200m} & \multicolumn{2}{c}{250m} & \multicolumn{2}{c}{300m}\\
    & AP & R@1 & AP & R@1 & AP & R@1 & AP & R@1 & AP & R@1 & AP & R@1 & AP & R@1 & AP & R@1\\
    \midrule
    LPN             & 63.50 & 58.20 & 74.16 & 69.60 & 79.70 & 75.60 & 82.93 & 78.50 & 63.68 & 77.50 & 78.36 & 87.50 & 84.26 & 90.00 & 87.99 & 92.50 \\
    LPN$\dagger$    & \textbf{64.24} & \textbf{59.77} & \textbf{74.39} & \textbf{70.38} & \textbf{81.14} & \textbf{77.50} & \textbf{84.26} & \textbf{79.38} & \textbf{64.88} & \textbf{78.40} & \textbf{80.13} & \textbf{88.74} & \textbf{85.50} & \textbf{93.75} & \textbf{89.72} & \textbf{93.41} \\
    \hline
    FSRA            & 82.69 & 78.70 & 88.66 & 85.65 & 91.26 & 88.95 & 93.40 & 91.50 & 83.65 & 93.75 & 90.01 & 93.75 & 91.67 & 97.50 & 92.49 & 95.00 \\
    {FSRA}$\dagger$   & \textbf{83.95} & \textbf{79.75} & \textbf{89.36} & \textbf{86.28} & \textbf{91.67} & \textbf{89.16} & \textbf{94.23} & \textbf{92.17} & \textbf{84.44} & \textbf{95.49} & \textbf{90.55} & \textbf{94.87} & \textbf{92.65} & \textbf{97.74} & \textbf{93.36} & \textbf{96.13} \\
    \hline
    SDPL            & 76.64 & 72.07 & 84.98 & 81.92 & 89.53 & 87.05 & 92.34 & 90.35 & 70.28 & 80.00 & 80.57 & 86.25 & 85.64 & 88.75 & 87.43 & 90.00 \\
    SDPL$\dagger$   & \textbf{77.11} & \textbf{75.30} & \textbf{85.19} & \textbf{83.00} & \textbf{91.25} & \textbf{88.99} & \textbf{92.97} & \textbf{90.86} & \textbf{71.43} & \textbf{80.80} & \textbf{81.84} & \textbf{87.10} & \textbf{85.81} & \textbf{91.25} & \textbf{88.52} & \textbf{91.25} \\
    \hline
    MCCG            & 81.21 & 79.96 & 86.24 & 85.01 & 92.15 & 90.47 & 94.97 & 94.20 & 89.76 & 92.06 & 92.40 & 93.88 & 96.15 & 96.34 & 96.52 & 98.78 \\
    MCCG $\dagger$  & \textbf{82.10} & \textbf{80.56} & \textbf{87.16} & \textbf{86.40} & \textbf{92.98} & \textbf{91.08} & \textbf{95.37} & \textbf{94.84} & \textbf{90.92} & \textbf{92.96} & \textbf{93.15} & \textbf{94.44} & \textbf{96.43} & \textbf{97.06} & \textbf{96.74} & \textbf{98.99} \\
    \hline
    Sample4Geo          & 96.08 & 94.75 & 97.69 & 96.75 & 98.38 & 97.25 & 98.41 & 97.20 & 95.60 & 96.25 & 96.41 & 96.25 & 96.54 & 96.25 & 96.57 & 97.50 \\
    Sample4Geo$\dagger$ & \textbf{97.12} & \textbf{94.97} & \textbf{98.05} & \textbf{97.19} & \textbf{98.63} & \textbf{98.00} & \textbf{98.99} & \textbf{98.34} & \textbf{96.24} & \textbf{96.87} & \textbf{96.90} & \textbf{97.22} & \textbf{96.98} & \textbf{98.01} & \textbf{97.11} & \textbf{97.82}\\
    \hline
    Game4Loc          & 95.59 & 94.62 & 97.27 & 96.55 & \textbf{98.16} & 97.55 & 98.24 & 97.67 & 93.06 & 93.75 & 94.50 & 96.25 & 94.92 & \textbf{96.25} & 95.36 & 95.00 \\
    Game4Loc$\dagger$ & \textbf{96.70} & \textbf{95.80} & \textbf{97.78} & \textbf{97.10} & 98.14 & \textbf{97.60} & \textbf{98.98} & \textbf{98.65} & \textbf{93.37} & \textbf{96.25} & \textbf{95.03} & \textbf{97.50} & \textbf{95.95} & \textbf{96.25} & \textbf{96.28} & \textbf{97.50} \\ 
    \bottomrule
\end{tabular}}
\vspace{-3mm}
\end{table*}

\subsection{Main Results}
\noindent \textbf{Results on University-1652.}
To evaluate the effectiveness of CVD, we integrate it into five representative baselines spanning diverse architectures, including ResNet, ConvNeXt, and Vision Transformer (ViT), and conduct experiments on University-1652.
As summarized in \cref{tab:univ1652}, CVD consistently improves performance across all backbones and evaluation metrics.
For instance, incorporating CVD into MCCG yields a \textbf{+1.75\%} improvement in R@1 (Drone$\rightarrow$Satellite), while Game4Loc shows notable gains of \textbf{+1.39\%} in AP and \textbf{+1.62\%} in R@1.
Note that CVD is used only during training and introduces no additional inference overhead.
These results support our hypothesis that explicitly disentangling \textit{content} and \textit{viewpoint} leads to more robust and discriminative representations for DVGL.

\noindent \textbf{Results on SUES-200.}
We evaluate CVD on SUES-200 to examine its robustness under varying levels of viewpoint disparity induced by different drone altitudes.
As reported in \cref{tab:sues200}, CVD improves the performance of all baselines across both matching directions and all altitude levels.
Notably, the relative gains are more pronounced at lower altitudes (\eg, 150m), where off-nadir distortions are most severe.
At this height, FSRA and LPN improve by \textbf{+1.46\%} and \textbf{+1.20\%} in AP (Satellite$\rightarrow$Drone), while SDPL sees a \textbf{+2.69\%} gain in R@1 (Drone$\rightarrow$Satellite).
As altitude increases and the viewpoint gap narrows, CVD continues to yield consistent improvements.
For instance, MCCG achieves \textbf{+0.40\%} gains in AP at 300m.
These results demonstrate that CVD significantly enhances cross-view matching robustness under different viewpoints and altitudes, particularly in low-altitude settings where viewpoint-induced distortions are most challenging.

\begin{table*}[!t]
\centering
\caption{Cross-dataset generalization results comparing baselines and their CVD-enhanced counterparts (marked with $\dagger$), trained on the University-1652 dataset and directly tested on the SUES-200 dataset.}
\label{tab:generalization}
\vspace{-3mm}
\setlength{\tabcolsep}{4pt}
\resizebox{\linewidth}{!}{
\begin{tabular}{lcccccccccccccccc}
    \toprule
    \multirow{3}{*}{Method} & \multicolumn{8}{c}{\textbf{Drone $\rightarrow$ Satellite}} & \multicolumn{8}{c}{\textbf{Satellite $\rightarrow$ Drone}} \\
    \cmidrule(lr){2-9} \cmidrule(lr){10-17}
    & \multicolumn{2}{c}{150m} & \multicolumn{2}{c}{200m} & \multicolumn{2}{c}{250m} & \multicolumn{2}{c}{300m} & \multicolumn{2}{c}{150m} & \multicolumn{2}{c}{200m} & \multicolumn{2}{c}{250m} & \multicolumn{2}{c}{300m}\\
    & AP & R@1 & AP & R@1 & AP & R@1 & AP & R@1 & AP & R@1 & AP & R@1 & AP & R@1 & AP & R@1\\
    \midrule
    LPN             & 42.83 & 36.70 & 52.99 & 46.72 & 59.42 & 53.62 & 62.15 & 56.55 & 25.30 & 30.00 & 34.36 & 38.75 & 38.53 & 42.50 & 43.92 & 53.75 \\
    LPN $\dagger$   & \textbf{44.93} & \textbf{38.58} & \textbf{56.68} & \textbf{50.50} & \textbf{62.16} & \textbf{55.73} & \textbf{64.89} & \textbf{58.60} & \textbf{29.21} & \textbf{32.50} & \textbf{39.18} & \textbf{42.50} & \textbf{45.98} & \textbf{52.50} & \textbf{52.52} & \textbf{60.00} \\
    \hline
    FSRA                 & 58.22 & 52.45 & 67.10 & 61.87 & 70.64 & 66.07 & 71.99 & 67.50 & 50.95 & 58.75 & 59.07 & 66.25 & 61.07 & 62.50 & 61.98 & 63.75 \\
    {FSRA}$\dagger$        & \textbf{62.41} & \textbf{56.43} & \textbf{70.11} & \textbf{65.47} & \textbf{74.06} & \textbf{70.08} & \textbf{75.27} & \textbf{71.42} & \textbf{53.08} & \textbf{60.71} & \textbf{61.46} & \textbf{68.01} & \textbf{64.47} & \textbf{67.43} & \textbf{65.48} & \textbf{67.75} \\
    \hline
    SDPL                  & 38.52 & 32.80 & 47.30 & 41.33 & 52.38 & 46.72 & 53.62 & 48.50 & 25.26 & 27.50 & 35.05 & 37.50 & 41.74 & 46.25 & 43.85 & 48.75 \\
    SDPL$\dagger$         & \textbf{40.14} & \textbf{35.59} & \textbf{50.82} & \textbf{45.72} & \textbf{55.63} & \textbf{50.12} & \textbf{56.38} & \textbf{51.00} & \textbf{28.36} & \textbf{31.67} & \textbf{37.99} & \textbf{40.48} & \textbf{45.04} & \textbf{51.55} & \textbf{50.79} & \textbf{58.23} \\
    \hline
    MCCG                  & 74.99 & 70.85 & 86.04 & 83.20 & 90.84 & 88.90 & 93.38 & 91.85 & 59.85 & 63.75 & 74.65 & 81.25 & 79.87 & 83.75 & 81.17 & 86.25 \\
    MCCG$\dagger$         & \textbf{76.32} & \textbf{73.51} & \textbf{88.39} & \textbf{85.89} & \textbf{92.64} & \textbf{90.78} & \textbf{95.30} & \textbf{94.44} & \textbf{62.35} & \textbf{67.24} & \textbf{77.94} & \textbf{83.25} & \textbf{82.30} & \textbf{85.87} & \textbf{83.38} & \textbf{89.00} \\
    \hline
    Sample4Geo            & 64.24 & 57.62 & 74.45 & 69.00 & 81.22 & 77.02 & 85.48 & 81.90 & 83.85 & 88.75 & 90.09 & 92.50 & 91.68 & 96.25 & 93.51 & 95.00 \\
    Sample4Geo$\dagger$   & \textbf{66.21} & \textbf{59.80} & \textbf{76.43} & \textbf{72.11} & \textbf{83.58} & \textbf{79.85} & \textbf{88.20} & \textbf{84.07} & \textbf{85.20} & \textbf{89.99} & \textbf{91.47} & \textbf{93.63} & \textbf{92.49} & \textbf{97.13} & \textbf{95.00} & \textbf{95.36} \\
    \hline
    Game4Loc           & 82.39 & 78.85 & 88.57 & 86.10 & 90.31 & 88.17 & 90.94 & 88.75 & 75.29 & 80.00 & 81.31 & 88.75 & 84.31 & 88.75 & 86.40 & 92.50 \\
    Game4Loc$\dagger$  & \textbf{86.12} & \textbf{82.87} & \textbf{91.77} & \textbf{89.70} & \textbf{93.59} & \textbf{91.92} & \textbf{94.44} & \textbf{92.92} & \textbf{75.37} & \textbf{87.50} & \textbf{84.25} & \textbf{90.00} & \textbf{87.97} & \textbf{95.00} & \textbf{90.21} & \textbf{95.00} \\
    \bottomrule
\end{tabular}}
\vspace{-5mm}
\end{table*}

\begin{table}[!t]
\centering
\caption{Comparison of baselines and their CVD-enhanced counterparts (marked with $\dagger$) on the CVUSA and CVACT datasets.}
\label{tab:cvusa_cvact}
\vspace{-3mm}
\setlength{\tabcolsep}{2pt}
\resizebox{\linewidth}{!}{
\begin{tabular}{lcccccccccccc}
    \toprule
    \multirow{2}{*}{Method} & \multicolumn{4}{c}{\textbf{CVUSA}} & \multicolumn{4}{c}{\textbf{CVACT\_val}} \\
    \cmidrule(lr){2-5} \cmidrule(lr){6-9}
    & R@1 & R@5 & R@10 & R@1\% & R@1 & R@5 & R@10 & R@1\% \\
    \midrule
    LPN                  & 85.43 & 95.20 & 96.91 & 99.40 & 78.86 & 89.97 & 92.07 & 95.34 \\
    LPN$\dagger$         & \textbf{87.15} & \textbf{96.43} & \textbf{97.26} & \textbf{99.49} & \textbf{80.08} & \textbf{91.10} & \textbf{93.16} & \textbf{96.69} \\
    \hline
    TransGeo             & 93.72 & 98.01 & 98.56 & 99.78 & 83.99 & 93.49 & 95.17 & 97.80 \\
    TransGeo$\dagger$    & \textbf{94.35} & \textbf{98.83} & \textbf{99.07} & \textbf{99.80} & \textbf{84.96} & \textbf{94.61} & \textbf{95.90} & \textbf{98.44} \\
    \hline
    GeoDTR               & 93.05 & 98.01 & 98.94 & 99.80 & 85.11 & 94.00 & 95.47 & 98.03 \\
    GeoDTR$\dagger$      & \textbf{93.92} & \textbf{98.70} & \textbf{99.26} & \textbf{99.83} & \textbf{85.72} & \textbf{95.19} & \textbf{96.63} & \textbf{98.74} \\
    \hline
    Game4Loc             & 98.12 & 99.08 & 99.46 & 99.83 & 90.19 & 96.00 & 97.12 & 98.60 \\
    Game4Loc$\dagger$    & \textbf{98.43} & \textbf{99.32} & \textbf{99.64} & \textbf{99.90} & \textbf{90.54} & \textbf{96.68} & \textbf{97.30} & \textbf{98.72} \\
    \hline
    Sample4Geo           & 98.43 & 99.15 & 99.42 & 99.81 & 90.29 & 95.98 & 97.04 & 98.53 \\
    Sample4Geo$\dagger$  & \textbf{98.67} & \textbf{99.40} & \textbf{99.78} & \textbf{99.89} & \textbf{90.94} & \textbf{96.80} & \textbf{97.51} & \textbf{98.81} \\
    \bottomrule
\end{tabular}}
\vspace{-5mm}
\end{table}

\noindent \textbf{Results on University-1652 $\rightarrow$ SUES-200.}
To evaluate the generalization ability of CVD to unseen scenes, we train models on University-1652 and directly evaluate them on SUES-200 without any fine-tuning.
As shown in \cref{tab:generalization}, CVD significantly improves performance across all metrics for both LPN (ResNet-50) and Game4Loc (ViT-B).
For example, it boosts R@1 by up to \textbf{+6.25\%} for LPN and \textbf{+7.5\%} for Game4Loc, even without access to the target domain during training.
Remarkably, these cross-dataset improvements exceed those typically obtained via in-domain training on SUES-200, which yields only improvements of \textbf{1-3\%}.
These results highlight that disentangling \textit{content} and \textit{viewpoint} leads to more transferable representations that generalize effectively across regions and views.

\noindent \textbf{Results on CVUSA and CVACT.}
To further examine the generality of CVD beyond the drone-view setting, we also evaluate it on CVUSA and CVACT, which involve ground-to-satellite matching under extreme viewpoint differences and substantial scene layout variation.
As shown in \cref{tab:cvusa_cvact}, all baselines exhibit continuous improvements when trained with CVD, even though they already perform strongly.
These improvements are particularly meaningful given the challenges posed by ground-view imagery, such as occlusion, illumination changes, and perspective distortion. 
By explicitly separating view-specific conflicting information, CVD enhances the consistency of content representations across modalities.
For example, Sample4Geo improves by \textbf{+0.24\%} and \textbf{+0.65\%} in R@1 on CVUSA and CVACT, respectively, while GeoDTR achieves \textbf{+0.77\%} and \textbf{+0.61\%} gains in R@10.

\noindent \textbf{Qualitative Results.}
Appendix Sec. \textcolor{cvprblue}{10.1}, we provide qualitative comparisons of cross-view retrieval results on University-1652 using both CNN-based LPN and Transformer-based Game4Loc. 
The results intuitively confirm that CVD disentangles view-agnostic content from viewpoint-specific variations, enabling more reliable cross-view correspondence.

\noindent \textbf{Training and Inference Time.} We report training and inference time, with comparisons to multiple baselines, in Appendix Sec. \textcolor{cvprblue}{9.1}. The CVD factorization halves the channel width (from $C$ to $C/2$), directly reducing the cost of contrastive similarity computation and retrieval indexing, thereby yielding faster training and inference speeds.
For example, LPN$\dagger$ reduces training time by 26\%, while Sample4Geo$\dagger$ reduces inference latency by 69\%.

\begin{figure*}[!t]
\centering
\includegraphics[width=\linewidth]{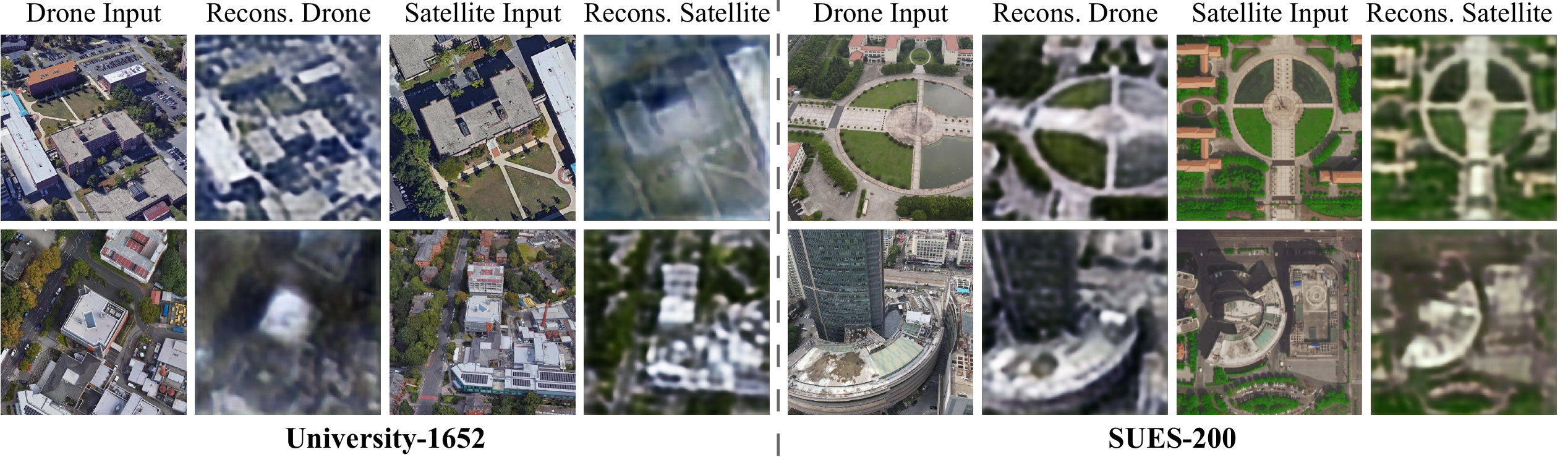} 
\vspace{-6mm}
\caption{Qualitative results of cross-view reconstruction on the University-1652 and SUES-200 datasets.}
\label{fig:reconstruct_quality}
\vspace{-3mm}
\end{figure*}

\begin{table*}[!t]
\caption{Ablation studies on:
(a) Effect of CVD's components.
(b) Content-viewpoint split ratio $\alpha$.
(c) Analysis of different $\tau$ in InfoNCE.}
\vspace{-4mm}
\begin{subtable}{.58\linewidth}
\centering 
\hfill
\caption{}
\resizebox{\linewidth}{!}{
\begin{tabular}{c|l|cc|cc}
    \specialrule{.09em}{.05em}{.05em} 
    \multirow{2}{*}{Exp.} & \multirow{2}{*}{Methods (ResNet50)} & \multicolumn{2}{c|}{\textbf{Drone $\rightarrow$ Satellite}} & \multicolumn{2}{c}{\textbf{Satellite $\rightarrow$ Drone}} \\
    & & AP & R@1 & AP & R@1 \\
    \hline 
    1 & Baseline LPN                         & 77.26 & 73.87 & 73.55 & 85.28  \\
    2 & w/o two constraints                  & 74.62 & 71.83 & 69.85 & 83.49 \\
    3 & Exp.1 + $\mathcal{L}_{\text{iic}}$   & 77.95 & 73.95 & 73.97 & 85.24 \\
    4 & Exp.1 + $\mathcal{L}_{\text{irc}}$   & 78.28 & 74.62 & 74.63 & 85.45 \\
    5 & \textbf{CVD} (LPN)                   & \textbf{78.99} & \textbf{75.78} & \textbf{74.89} & \textbf{85.88}  \\
    \hline
    6 & Baseline Game4Loc                    & 72.90 & 68.49 & 66.53 & 83.16 \\
    7& w/o two constraints                   & 69.71 & 65.85 & 64.51 & 81.19 \\
    8 & Exp.6 + $\mathcal{L}_{\text{iic}}$   & 73.29 & 69.19 & 66.70 & 83.20 \\    
    9 & Exp.7 + $\mathcal{L}_{\text{irc}}$   & 73.60 & 69.78 & 67.03 & 83.61 \\
    10& \textbf{CVD} (Game4Loc)              & \textbf{74.31} & \textbf{70.46} & \textbf{67.60} & \textbf{83.67}  \\
    \specialrule{.09em}{.05em}{.05em} 
\end{tabular}}
\label{tab:ablation_components}
\end{subtable}
\begin{subtable}{.4\linewidth}
\centering 
\hfill
\caption{}
\resizebox{\textwidth}{!}{
\begin{tabular}{c|cc|cc}
\specialrule{.1em}{.05em}{.05em} 
    \multirow{2}{*}{Split Ratio} & \multicolumn{2}{c|}{\textbf{Drone $\rightarrow$ Satellite}} & \multicolumn{2}{c}{\textbf{Satellite $\rightarrow$ Drone}} \\
    & AP & R@1 & AP & R@1 \\
    \hline
    $\alpha=1/3$  & 78.13 & 74.90 & 74.08 & 84.31 \\
    $\alpha=1/2$  & \textbf{78.99} & \textbf{75.78} & \textbf{74.89} & \textbf{85.88}  \\
    $\alpha=3/4$  & 78.15 & 74.76 & 73.28 & 86.59 \\
    \hline
    No squeeze    & 78.06 & 74.85 & 74.48 & 84.85 \\
\specialrule{.1em}{.05em}{.05em} 
\end{tabular}}
\label{tab:ablation_ratio}
\vspace{-2.1mm}
\centering 
\hfill
\caption{}
\resizebox{\textwidth}{!}{
\begin{tabular}{l|cc|cc}
    \specialrule{.1em}{.05em}{.05em} 
    \multirow{2}{*}{Method} & \multicolumn{2}{c|}{\textbf{Drone $\rightarrow$ Satellite}} & \multicolumn{2}{c}{\textbf{Satellite $\rightarrow$ Drone}} \\
    & AP & R@1 & AP & R@1 \\
    \hline
    $\tau=0.07$ & \textbf{78.89} & 75.74 & 74.28 & 85.03 \\
    $\tau=0.05$ & \textbf{78.99} & \textbf{75.78} & \textbf{74.89} & \textbf{85.88} \\
    $\tau=0.03$ & 78.98 & 75.76 & 74.61 & 85.43 \\
    Bi-InfoNCE  & 78.72 & 75.14 & 74.36 & 85.21 \\
\specialrule{.1em}{.05em}{.05em} 
\end{tabular}}
\label{tab:ablation_tau}
\end{subtable}
\vspace{-3mm}
\end{table*}

\begin{table*}[!t]
\centering
\caption{Effect of different reconstruction loss functions on the University-1652 dataset.}
\vspace{-3mm}
\label{tab:reconstruct_loss}
\setlength{\tabcolsep}{8pt}
\resizebox{\textwidth}{!}{
\begin{tabular}{lcccccccccccc}
    \toprule
    \multirow{2}{*}{Loss} & \multirow{2}{*}{PSNR$\uparrow$} & \multirow{2}{*}{SSIM$\uparrow$} & \multicolumn{5}{c}{\textbf{Drone $\rightarrow$ Satellite}} & \multicolumn{5}{c}{\textbf{Satellite $\rightarrow$ Drone}} \\
    \cmidrule(lr){4-8} \cmidrule(lr){9-13}
    & & & AP & R@1 & R@5 & R@10 & R@1\% & AP & R@1 & R@5 & R@10 & R@1\%\\
    \midrule
    MSE         & \textbf{19.38} & \textbf{0.4608} & \textbf{93.95} & \textbf{92.94} & \textbf{97.59} & \textbf{98.28} & \textbf{98.32} & \textbf{91.92} & \textbf{94.86} & 96.71 & 97.14 & \textbf{99.57} \\
    SSIM        & 19.21 & 0.4601 & 93.19 & 92.05 & 97.26 & 97.87 & 97.90 & 91.78 & 94.55 & 96.40 & 96.19 & 99.56 \\
    Perceptual  & 19.06 & 0.4589 & 92.70 & 91.45 & 96.82 & 97.84 & 97.91 & 90.85 & 94.14 & \textbf{97.42} & \textbf{97.26} & 99.29\\
    \bottomrule
\end{tabular}}
\vspace{-6mm}
\end{table*}

\subsection{Effectiveness of Disentangled Strategy}
To evaluate our disentanglement strategy, we conduct cross-view image reconstruction (Drone $\rightarrow$ Satellite) on University-1652 and SUES-200.
As illustrated in \cref{fig:reconstruct_quality}, the visualization results show that 
the reconstructed outputs consistently preserve the global layout, semantic topology, and structural relations of the original scenes, despite the loss of certain high-frequency and color details.
For example, the upper-right sample faithfully recovers the circular central plaza and the relative arrangement of surrounding buildings.
This indicates that CVD successfully learned meaningful \textit{content} and \textit{viewpoint} representations, thereby verifying the effectiveness of disentangling.

\noindent\textbf{Note.} Additional visualizations are provided in Appendix Secs. \textcolor{cvprblue}{10.2} to \textcolor{cvprblue}{10.4}, including more cross-view reconstructions and attention maps.

\subsection{Ablation Study}
\noindent \textbf{Effect of CVD's Components.}
We conduct ablation studies on the University-1652 dataset to evaluate the contribution of each component in CVD, as summarized in \cref{tab:ablation_components}.
To ensure fair comparison, we adopt two representative pipelines, LPN and Game4Loc, which both use a shared ResNet50 backbone.
Removing both constraints (Exp.2 and 7) leads to a notable performance drop, while individually adding the intra-view independence (Exp.3 and 8) or inter-view reconstruction constraint (Exp.4 and 9) yields consistent gains.
The best results are obtained when both constraints are jointly applied (Exp.5 and 10), confirming that explicitly factorizing \textit{content} and \textit{viewpoint} is essential for robust cross-view alignment.

\noindent \textbf{Different Content-Viewpoint Ratio.}
We investigate the impact of different split ratios between content embedding and viewpoint embedding, as reported in \cref{tab:ablation_ratio}.
Assigning an imbalanced proportion of dimensions, favoring either content ($\alpha=3/4$) or viewpoint ($\alpha=1/3$), results in performance drops of 0.84\% and 0.81\% in AP (Drone$\rightarrow$Satellite), respectively.
The best performance occurs when $\alpha=1/2$, indicating that balanced factorization most effectively preserves factor-specific information.
Interestingly, the “No squeeze” setting, where both branches retain full dimensionality, also underperforms the balanced configuration by 0.93\% in AP, suggesting that moderate compression encourages more effective disentanglement.

\noindent \textbf{Analysis of Temperature Parameters.}
As illustrated in \cref{tab:ablation_tau}, model performance remains stable across a range of temperature values $\tau$.
This insensitivity indicates that the performance gains primarily result from effective representation disentanglement rather than contrastive loss tuning.

\noindent \textbf{Effect of Reconstruction Losses.}
We compare MSE, SSIM, and perceptual losses. As shown in \cref{tab:reconstruct_loss}, MSE achieves the highest PSNR/SSIM and best retrieval accuracy, indicating that pixel-level fidelity better preserves view consistency.
By contrast, SSIM and perceptual losses relax strict photometric fidelity and show higher tolerance to small misalignments, which may reduce geometric consistency and slightly affect retrieval performance. We therefore adopt MSE in all experiments.

\noindent \textbf{Note.} We provide additional ablations in the Appendix, including robustness to training data scale (Tab. \textcolor{cvprblue}{8}), the number of SWD projections (Tab. \textcolor{cvprblue}{9}), and the effect of loss-balancing weights (Tab. \textcolor{cvprblue}{11}), among others.
\section{Conclusion}
In this paper, we revisit drone-view geo-localization (DVGL) from a manifold-learning perspective and propose \textbf{CVD}, a unified framework that explicitly disentangles \textit{content} and \textit{viewpoint} within visual representations. 
CVD follows an embed-disentangle-reconstruct paradigm, guided by intra-view independence and inter-view reconstruction constraints, to promote \textit{content}- and \textit{viewpoint}-specific encoding.
Extensive experiments show that CVD consistently improves cross-view matching accuracy across diverse pipelines and enhances robustness and generalization under varied scenarios, viewpoints, and altitudes, while achieving lower inference latency.
These results underscore the value of separating \textit{content} and \textit{viewpoint} for DVGL and point toward more robust and generalizable DVGL systems.
We discuss limitations and directions for future work in the Appendix.
{
    \small
    \bibliographystyle{ieeenat_fullname}
    \bibliography{main}
}


\end{document}